\documentclass[11pt,a4paper]{article}
 
\usepackage[margin=1in]{geometry}
\usepackage[T1]{fontenc}
\usepackage{lmodern}
\usepackage{graphicx}
\usepackage{amsmath}
\usepackage{amssymb}
\usepackage{amsthm}
\usepackage{makecell}
\usepackage{natbib}
\usepackage{booktabs}
\usepackage{comment}
\usepackage{multirow}
\usepackage{color}
\usepackage{url}
\usepackage{subcaption}
\usepackage[colorlinks=true,linkcolor=blue,citecolor=blue,urlcolor=blue]{hyperref}
 
\newcommand{\D}{D}
\newcommand{\Dtr}{D_{\text {tr}}}

\newcommand{\free}{\mathcal{M}}  
\newcommand{\So}{\mathcal{S}_{\text {o}}}
\newcommand{\bsy}{q}

\newcommand{\dmodel}{{d_{\rm model}}}
\newcommand{\nin}{c}
\newcommand{\N}{n}
\newcommand{\R}{\mathbb{R}}

\newcommand{\G}{\mathbb{G}}

\theoremstyle{remark}
\newtheorem{remark}{Remark}

\begin{document}
 
\title{In-Context Learning for System Identification: Probabilistic Formulation, Long-Sequence
Handling, and Benchmark Validation\thanks{This work was partly supported by the Hasler Foundation project ``FALCON - Foundation models for zero-shot adaptive control and prediction for dynamical systems'' and as a part of NCCR Automation, a National Centre of Competence in Research, funded by the Swiss National Science Foundation (grant number 51NF40\_225155).}}
 
\author{
Matteo Rufolo$^{1,2}$ \and Dario Piga$^{1}$ \and Filippo Pura$^{1}$ \and Marco Forgione$^{1}$ \\[1.5ex]
\small $^{1}$SUPSI, IDSIA - Dalle Molle Institute for Artificial Intelligence, Lugano, Switzerland \\
\small $^{2}$USI, Faculty of Informatics, Lugano, Switzerland \\[1ex]
\small \texttt{\{matteo.rufolo, dario.piga, filippo.pura, marco.forgione\}@supsi.ch}
}
\date{}
 
\maketitle

\noindent\rule{\textwidth}{1pt}
Please cite this version of the paper:\\ \\
M. Rufolo, D. Piga, F. Pura and M. Forgione  ``In-Context Learning for System Identification: Probabilistic Formulation, Long-Sequence Handling, and Benchmark Validation," in \textit{Automatica},
2026. In press.
\vskip 1em \noindent
You may use the following bibtex entry:
\begin{verbatim}
@article{rufolo2026incontext,
  title={In-Context Learning for System Identification: Probabilistic Formulation,
  Long-Sequence Handling, and Benchmark Validation},
  author={Rufolo, Matteo and  Piga, Dario and Pura, Filippo and Forgione, Marco},
    journal={Automatica},
  year={2026},
  note={in press}
}
\end{verbatim}
\noindent\rule{\textwidth}{1pt} 

\begin{abstract}
Traditional system identification typically involves two  steps: initially, training data is processed to estimate a model of the system of interest; then, the  model is used to predict outputs for new input queries. 
In contrast, recent in-context learning approaches learn a meta model that maps directly from training data and query inputs to outputs in a zero-shot fashion, bypassing explicit model estimation. This approach leverages a meta model trained offline, on large sets of synthetically generated data, to emulate the behavior of a system identification algorithm. In this work, we advance this meta-learning framework for system identification in three directions. First, we develop a unified probabilistic meta-identification architecture that combines a Transformer-based encoder–decoder with tailored mechanisms for handling long training sequences, enabling calibrated predictions over long horizons. Second, we show how a pre-trained meta model can act as an uncertainty-aware synthetic data generator for classical system identification, providing task-specific data that regularize parametric models when real data are scarce. Third, we systematically evaluate the framework on standard nonlinear system identification benchmarks and on a challenging out-of-distribution scenario, demonstrating competitive accuracy with state-of-the-art methods and strong generalization, including cases where input-only adaptation is sufficient to handle significant distribution shifts.
\end{abstract}


\section{Introduction}
\subsection{System identification and meta learning}
System identification (SYSID) provides data-driven methods to construct mathematical models of dynamical systems from measured input–output data. Given a dataset, the goal is to estimate a model that captures the underlying behavior with sufficient accuracy for tasks such as simulation, prediction, and control design. Over the past decades, a broad range of identification techniques has been developed, ranging from prediction-error and subspace methods for linear to deep-learning-based approaches for nonlinear model learning; see, e.g.,~\cite{forgione2021continuous,zhou2022sparse,beintema2021nonlinear,masti2021learning,schoukens2014identification,paduart2010identification}. 
In the classical setting, each identification task is posed for a single system, and a model is fitted using exclusively the data available for that system.

\paragraph*{Traditional identification limitations}
In standard practice, the estimation pipeline relies heavily on expert intervention. The practitioner must select a suitable model structure (e.g., linear \emph{vs} nonlinear, time-invariant \emph{vs} time-varying, state-space \emph{vs} input/output representation, etc.), choose an identification approach (e.g., prediction error methods, subspace methods, kernel-based approaches, etc.), and tune hyper-parameters in the numerical optimization routines used to solve the identification problem (e.g., learning rate in gradient-based methods, regularization weights, stopping criteria to prevent overfitting, etc.). These design and tuning choices are typically repeated from scratch for each new system of interest.

In many applications, however, the goal is not to model a single isolated plant, but to repeatedly identify 
related systems (for example, members of a given benchmark family or class of processes). In such settings, traditional pipelines treat each system independently: for every new dataset, the modeling and tuning steps described above are repeated, and a separate model is estimated from scratch. This per-system workflow does not exploit information gained from previously identified systems and can be computationally demanding when many systems must be modeled. Moreover, classical approaches typically yield a different model for each system, which complicates downstream tasks such as deploying a shared controller across a fleet of systems or transferring prior knowledge to new tasks. While uncertainty-aware identification methods and advanced long-sequence models exist, they are not generally designed to provide class-level, zero-shot predictions for new systems from a short context trajectory.

\paragraph*{Proposed meta-learning approach}
To address these challenges, we adopt a meta-learning viewpoint in which the basic object of interest is a \emph{class} of dynamical systems rather than a single plant. Building on the in-context system identification paradigm introduced in~\cite{forgione2023context},  this paper presents a \emph{novel} SYSID paradigm inspired by meta learning techniques~\cite{hospedales2021meta}. We consider a meta model that is trained once, offline, on a large synthetic corpus generated from a distribution over systems. At inference time, this pre-trained meta model receives a context input–output sequence from a new system and a query input, and directly predicts the corresponding output in a zero-shot fashion eliminating the need to implement any system identification algorithm and avoiding the selection of model structure and hyper-parameters.

This class-level meta-modeling addresses the limitations discussed above in the following ways:
\begin{itemize}
    \item \emph{Reuse of knowledge across systems:} a single meta model is shared by all systems in the class, so information gained during offline training is amortized across tasks instead of being discarded after each identification.
    \item \emph{Reduced per-system design effort:} architecture and training hyper-parameters are chosen once at the class level. For a new system, the meta model is applied without revisiting model-structure selection or hyper-parameter tuning for each trajectory.
    \item \emph{Zero-shot prediction from short contexts:} given a limited context from a new system, the meta model can produce output predictions for new inputs without any task-specific gradient-based adaptation, in contrast to classical pipelines that require solving a new optimization problem per system.
\end{itemize}

This enables knowledge reuse across systems and reduces the need to repeatedly perform full identification procedures—including structure and hyper-parameter selection—for every new trajectory.

We emphasize that addressing the above limitations comes at its own cost. 
In particular, the proposed meta-learning paradigm relies on a substantial offline pre-training phase on synthetic data, and a performance gap with carefully tuned classical system identification methods may still appear when abundant task-specific data are available or when strict requirements on parsimonious models for control design are imposed.
These trade-offs, and the regimes in which the meta-modeling approach is most advantageous, are examined in detail in the subsequent sections.

The proposed meta learning approach is analogous to \emph{Large Language Models} (LLMs) like GPT-4, LLaMA,  Gemini, and DeepSeek, which are pre-trained on vast text corpora to understand and generate text. Similarly, our meta learning approach enables users to generate output predictions by providing context and a related query, leveraging knowledge acquired in pre-training to efficiently handle various system identification tasks.

\subsection{Related Literature} 
Meta learning, also referred to as learning to learn, is a branch of machine learning introduced in the late 1980s~\cite{schmidhuber1987evolutionary} that has gained significant momentum in recent years. In a broad sense, the goal of meta learning is to enhance model and algorithm efficiency by leveraging experience from multiple related tasks. While standard machine learning trains models for a single task, meta learning approaches aim to capture common patterns and similarities across tasks, enabling rapid adaptation to new ones with minimal data and computational resources.

One of the most popular meta learning algorithms is \emph{Model-Agnostic Meta Learning} (MAML)~\cite{finn2017model}, which aims to find model parameters that can quickly adapt to new tasks. MAML has proven particularly effective in few-shot learning scenarios across various domains and applications, including computer vision for few-shot image classification tasks~\cite{antoniou2019how},  reinforcement learning to accelerate policy learning in new environments~\cite{finn2017model}, materials science for predicting material properties and accelerating materials discovery~\cite{zuo2021accelerating}, as well as in system identification, optimization  and control~\cite{kaushik2020fast,chakrabarty2023meta,richards2021adaptive,chakrabarty2025meta}.

Another instance of meta learning is represented by in-context learning, where large models  
are trained on diverse and extensive datasets, enabling them to process a context--a demonstration of  the task at hand--along with a query input, and directly predict the corresponding output without further training~\cite{kirsch2022general, dong2024survey}. In this framework, zero-shot prediction emerges as a powerful feature of in-context learning, where models  leverage their extensive pre-training on diverse datasets to adapt   to new tasks without requiring additional fine-tuning or retraining. Typical application of in-context learning include Natural Language Processing~\cite{zhu2023multilingual}, vision~\cite{zhou2024visual}, speech recognition \cite{chen2024salm}, and robotics \cite{zhu2024incoro}, just to cite a few. In the field of system identification, in-context learning has recently demonstrated promising results for zero-shot prediction, leveraging knowledge gained from estimating the behavior of systems similar to the target one \cite{forgione2023context,rufolo2025distributionally,ZheHal23}.

\subsection{Paper Contribution}

{
This paper builds upon the preliminary work on in-context learning for system identification introduced in~\cite{forgione2023context} and further developed in~\cite{piga2024synthetic, rufolo2024enhanced}.
The aim is to provide a unified framework that integrates and extends these previous contributions, establishing a comprehensive and probabilistic formulation of in-context learning for system identification.
Within this unified setting, the present work introduces the following technical contributions:
}
 
\begin{itemize}
    \item {\textbf{Extended methods for long-context processing:} 
    A novel ensemble-based strategy with probabilistic weighting is proposed to efficiently handle long input–output context sequences. 
    Compared with the patching approach of~\cite{rufolo2024enhanced}, the new method achieves higher in-distribution accuracy and reduced meta-training time, at the cost of a slight increase in meta-inference time.
        } 
    \item {\textbf{Probabilistic synthetic data generation:} The data generation strategy of~\cite{piga2024synthetic} is reformulated within the probabilistic framework, enabling the production of uncertainty-aware synthetic datasets for training classical system identification algorithms.}
    \item {\textbf{Benchmark validation:}  Where~\cite{forgione2023context, rufolo2024enhanced} focused mainly on simulated scenarios, we systematically evaluate the full proposed architecture on multiple nonlinear system identification benchmarks from the nonlinear benchmark initiative~\cite{NonLinearBench}, demonstrating robustness and generalization on real experimental data.}
    \item {\textbf{Model adaptation study:} We assess the framework in an out-of-distribution (OOD) scenario in which both the dynamical systems and the input signals differ from those used during meta model training. We show that fine-tuning by matching only the inputs, while keeping the family of training systems fixed, achieves performance comparable to full adaptation. This result is practically relevant for scenarios where the system dynamics are uncertain but the excitation signals, chosen by the user, are known in advance.}
\end{itemize}

Taken together, these contributions go beyond our previous works~\cite{forgione2023context,piga2024synthetic,rufolo2024enhanced} in several ways. First, the probabilistic formulation of synthetic data generation yields uncertainty-aware synthetic datasets and a principled mechanism to down-weight unreliable synthetic samples, addressing the growing need for more reliable synthetic data in machine learning. Second, the adaptation study shows that fine-tuning only the input side of the meta model can already recover most of the benefits of full adaptation, which is particularly relevant in applications where excitation signals can be designed but the system dynamics remain uncertain. Finally, by combining these elements with long-context processing in a single architecture and by carrying out a systematic evaluation on standard nonlinear benchmarks, including systems that differ from the WH training class, the paper validates the sim-to-real capabilities of the in-context approach and provides some of the first concrete tools for using meta-learning in system identification, considering the emerging interested in this research direction.

The rest of the paper is organized as follows. In Section~\ref{sec:Framework}, we review the meta modeling framework originally introduced in~\cite{forgione2023context}. We first formulate a point estimation problem of system outputs as in~\cite{forgione2023context}, and then extend this formulation to the more general case of probabilistic predictions. Model architectures and training algorithms are discussed in Section~\ref{sec:long_seq}, followed by novel approaches to manage long context sequences. The first one is based on sequence patching and inspired by the Vision Transformer~\cite{dosovitskiy2021imagepatching} for image classification. The second one relies on an ensembling of the meta model predictions associated with different (short) context sequences. Section~\ref{sec:synthetic_data_generation} details how the pre-trained meta model can generate synthetic data useful for classical system identification to estimate a parametric model of a dynamical system of interest when the training data are scarce or insufficiently informative. Examples showcasing the effectiveness of the  meta learning paradigm are reported in Section~\ref{sec:results}{, including benchmark performance and model adaptation analyses.}  Concluding remarks are drawn in Section~\ref{sec:conclusions}, along with directions for future research.

{\section{Notation}
Vectors are denoted by lowercase letters (e.g., $u$). Subscripts indicate the time index (e.g., $u_i$). For compactness, $u_{a:b}$ denotes the sequence $\{u_a, u_{a+1}, \ldots, u_b\}$. A superscript in parentheses, such as $u^{(i)}$, denotes the $i$-th sample drawn from a dataset or distribution. A superscript without parentheses, such as $u^i$, denotes the $i$-th subsequence extracted from a longer sequence. The same notation applies to the corresponding model outputs associated with each subsequence or sample. $\mathbb{E}_p[\cdot]$ denotes the expectation with respect to a probability distribution $p$. The Euclidean-norm of a vector $u$ is denoted as $\Vert u \rVert$. Estimated quantities are indicated by a hat, e.g., $\hat{y}$ is an estimate of $y$. $\mathcal{N}(\cdot; \mu, \sigma^2)$ denotes a Gaussian probability density function with mean $\mu$ and variance $\sigma^2$.
}
\section{Meta learning framework}
\label{sec:Framework}

\subsection{System Class} 
\label{sec:system-class}
Let us consider a class $\mathcal{S}$ of dynamical systems. This class could correspond, for example, to Linear Time-Invariant (LTI) systems, fading memory nonlinear systems, or a more restricted subset of dynamics describing specific real-world systems. Examples of such restricted classes include the set of all robotic manipulators with a maximum number of degrees of freedom or the set of all vehicles with specific characteristics, to name just a few.

We assume to have access to a (potentially infinite) stream of input/output datasets generated by different systems $\{{S}^{(i)}, i = 1, 2, \dots\}$ extracted from a probability distribution $p(S)$ defined over the class $\mathcal{S}$. 
For each system $S^{(i)}$, a context dataset $(u_{1:m}^{(i)}, y_{1:m}^{(i)})$ and a query dataset $(\tilde{u}_{1:\N}^{(i)}, \tilde{y}_{1:\N}^{(i)})$ are available, with 
$u_k, \tilde u_k \in \R^{n_u}$, $y_k, \tilde y_k \in \R^{n_y}$.
These two datasets are obtained by processing two randomly generated input sequences $u_{1:m}^{(i)}, \tilde{u}_{1:\N}^{(i)}$ through the same system ${S}^{(i)}$, starting from different random initial conditions. 

The systems ${S}^{(i)}$ may be instances of a high-fidelity simulator of a process of interest, with physical parameter randomly distributed according to domain knowledge. 
Overall, we assume there is an underlying \emph{distribution} $p(\D)$ of datasets generated by the system distribution $p(S)$ and the random input distribution.  The distribution $p(\D)$ may not be explicitly known, but samples $\{\D^{(i)} = (u_{1:m}^{(i)}, y_{1:m}^{(i)}), (\tilde{u}_{1:\N}^{(i)}, \tilde{y}_{1:\N}^{(i)}), i = 1, 2, \dots\}$ can be drawn from $p(\D)$.

\subsection{Classic system identification \emph{vs} meta modeling}

Classic system identification aims to estimate a different model for each system ${S}^{(i)}$ based on the available dataset $\D^{(i)}$. A classic approach might require choosing a model structure and a learning algorithm, using the context data $(u_{1:m}^{(i)}, y_{1:m}^{(i)})$.   Then, the system state at a time step $\nin < \N$ of the query may be estimated from the input/output samples $(\tilde{u}_{1:\nin}^{(i)}, \tilde{y}_{1:\nin}^{(i)})$ up to time $\nin$. By simulating the identified system model starting from the estimated initial condition and for the remaining query input sequence $\tilde{u}_{\nin+1:\N}^{(i)}$, one could make predictions $\hat{y}_{\nin+1:\N}^{(i)}$ of the output $y_{\nin+1:\N}^{(i)}$:
\begin{equation} 
\label{eq:sysid_est_sim}
\hat{y}_{\nin+1:\N}^{(i)} = \free(\tilde{u}_{\nin+1:\N}^{(i)}, \tilde{u}_{1:\nin}^{(i)}, \tilde{y}_{1:\nin}^{(i)}, u_{1:m}^{(i)}, y_{1:m}^{(i)}),
\end{equation}
where $\free$ represents the combined identification, system state estimation, and simulation procedure applied to the dataset $\D^{(i)}$.

\vspace{0.5em}
In the meta learning approach discussed in this paper, a single meta model is estimated for all systems in the class $\mathcal{S}$. Given a context dataset generated by a specific system ${S}^{(i)}$ and a query input, the meta model directly provides an estimate of the corresponding query output. 
{Specifically, we introduce a (black-box) map $\free_{\phi}$, parameterized by $\phi$, which processes the same inputs as in~\eqref{eq:sysid_est_sim}:}
\begin{equation} 
\label{eq:metamodel} 
\hat{y}_{\nin+1:\N}^{(i)} = \free_{\phi}(\tilde{u}_{\nin+1:\N}^{(i)}, \tilde{u}_{1:\nin}^{(i)}, \tilde{y}_{1:\nin}^{(i)}, u_{1:m}^{(i)}, y_{1:m}^{(i)}).
\end{equation}
The map $\free_{\phi}$ is trained to behave as a meta model for the system class $\mathcal{S}$ by minimizing the expected simulation error over the dataset distribution $p(\D)$ in an end-to-end supervised manner, i.e.,
\begin{equation}
    \label{eq:simulation_objective_samples}
\hat{\phi} = \arg \min_{\phi} 
    \overbrace{
  \mathbb{E}_{p(\D)}
    \left [ 
    \lVert \tilde{y}_{\nin+1:\N} -  \free_{\phi}(X) \rVert^2
    \right]
}^{=J(\phi)},
\end{equation}
where $J(\phi)$ is the loss function, and $X=(\tilde{u}_{\nin+1:\N}, \tilde{u}_{1:\nin}, \\\tilde{y}_{1:\nin}, u_{1:m}, y_{1:m})$ is introduced to compact the notation and represents all the input variables to the meta model $\free_{\phi}$. 

\vspace{0.5em}
The expectation over $p(\D)$ is approximated by a Monte Carlo average over $b$ system instances from the class $\mathcal{S}$ and corresponding measured context $(u_{1:m}^{(i)}, y_{1:m}^{(i)})$ and query sequences $(\tilde{u}_{1:\N}^{(i)}, \tilde{y}_{1:\N}^{(i)})$, i.e.,
\begin{equation} 
\label{eq:J_point}
J(\phi) \approx \frac{1}{b}\sum_{i=1}^b
    \lVert \tilde{y}_{\nin+1:\N}^{(i)} -  \free_{\phi}(X^{(i)}) \rVert^2.
\end{equation}

\subsection{Probabilistic Prediction}
{The meta model in~\eqref{eq:metamodel} provides point estimates of the future outputs. 
While informative, these estimates lack the expressiveness of a full probabilistic description, which would also quantify predictive uncertainty. 
To address this limitation, the meta model $\free_{\phi}$ is modified to output, at each time step, the parameters of a probability distribution over future outputs.}

Training of the meta model $\free_{\phi}$ is then performed with the aim of approximating the unknown conditional probability distribution $p(\tilde{y}_{\nin+1:\N} | X)$ with a parameterized distribution $q_\phi(\tilde{y}_{\nin+1:\N} | X)$. To this aim, we introduce the Kullback-Leibler (KL) divergence $\text{KL}(p \parallel q_{\phi})$ between $p$ and $q_{\phi}$:
\begin{equation}
    \label{eq:KL}
    \text{KL}(p \parallel q_\phi) = \mathbb{E}_{p(\tilde{y}_{\nin+1:\N} | X)} \left[ \log \frac{p(\tilde{y}_{\nin+1:\N} | X)}{q_{\phi}(\tilde{y}_{\nin+1:\N} | X)} \right].
\end{equation}
{The parameters $\phi$ defining $\free_{\phi}$ (and hence $q_\phi$) are chosen to minimize the expected KL divergence over the conditioning variables $X$:}
\begin{multline}
    \label{eq:KL_simplified}
    \mathbb{E}_{p(X)} \big [ \text{KL}(p \parallel q_\phi) \big ] = \mathbb{E}_{p(\D)}  \left [  \log \frac{p(\tilde{y}_{\nin+1:\N} | X)} {q_{\phi}(\tilde{y}_{\nin+1:\N} | X)} \right] \\
    = \overbrace{\mathbb{E}_{p(\D)} [-\log q_{\phi}(\tilde{y}_{\nin+1:\N} | X) ]}^{J(\phi)} + K,
\end{multline}
where $J(\phi)$ is the loss function minimized with respect to $\phi$, and $K$ is a constant independent of $\phi$.

{As in the point-estimate formulation, the expected loss $J(\phi)$ is approximated by a Monte Carlo average:}
\begin{equation}
    \label{eq:J_prob}
    J(\phi) \approx \frac{1}{b} \sum_{i=1}^b -\log q_\phi(\tilde{y}_{\nin+1:\N}^{(i)} | X^{(i)}),
\end{equation}
where $\tilde{y}_{\nin+1:\N}^{(i)}$ and $X^{(i)}$ (with $i = 1, \dots, b$) are samples from $p(\D)$ associated with the system ${S}^{(i)}$.

\vspace{0.5em}
In this paper, we parameterize the distribution $q_\phi$ as a multivariate Gaussian with a diagonal covariance matrix. For notation simplicity, we present the case of single-output systems (i.e., $n_y = 1$). Thus, $q_\phi$ is factorized as:
\begin{subequations} \label{eq:fact_q}
\begin{align}
    q_\phi(\tilde{y}_{\nin+1:\N} | X) &= \prod_{k=\nin+1}^{\N} q_\phi(\tilde{y}_k | X),\label{prob_output} \\
    q_\phi(\tilde{y}_k | X) &= \mathcal{N}(\tilde y_k;\, \mu_k(X, \phi), \sigma_k(X,\phi)^2) \label{prob_output_sample} 
\end{align}
\end{subequations}
The vectors of means $\mu_{\nin+1:\N}$ and standard deviations $\sigma_{\nin+1:\N}$, both of dimension $\N-\nin$, are the outputs of the meta model $\free_\phi$ when fed by the query input $u_{\nin+1:\N}$, initial input/output measurements $\tilde{u}_{1:\nin}, \tilde{y}_{1:\nin}$, and context data $u_{1:m}, y_{1:m}$:
\begin{equation}
    \label{eq:Model_free_sim}
    \mu_{\nin+1:\N}, \sigma_{\nin+1:\N} = \free_\phi(\tilde{u}_{\nin+1:\N}, \tilde{u}_{1:\nin}, \tilde{y}_{1:\nin}, u_{1:m}, y_{1:m}).
\end{equation}
{Here, $\mu_{\nin+1:\N}$ corresponds to the previous point estimate $\hat{y}_{\nin+1:\N}$, while $\sigma_{\nin+1:\N}$ quantifies the associated predictive uncertainty.}

{Intuitively, the meta model $\free_{\phi}$ acts as a system identification algorithm that infers the dynamics of each system ${S}^{(i)}$ from its input/output context $(u_{1:m}^{(i)}, y_{1:m}^{(i)})$. 
Using this inferred knowledge, it generates predictions and corresponding uncertainty estimates $(\mu_{\nin+1:\N}, \sigma_{\nin+1:\N})$ for the query sequence. 
Once trained, $\free_\phi$ can produce predictions for previously unseen systems using only an input/output context and a query input $\tilde{u}_{\nin+1:\N}$, without the need to identify a dedicated model for each system.}

\subsection{Meta model architecture and training}

\begin{figure*}[!tb]
\centering
\includegraphics[width=.95\textwidth]{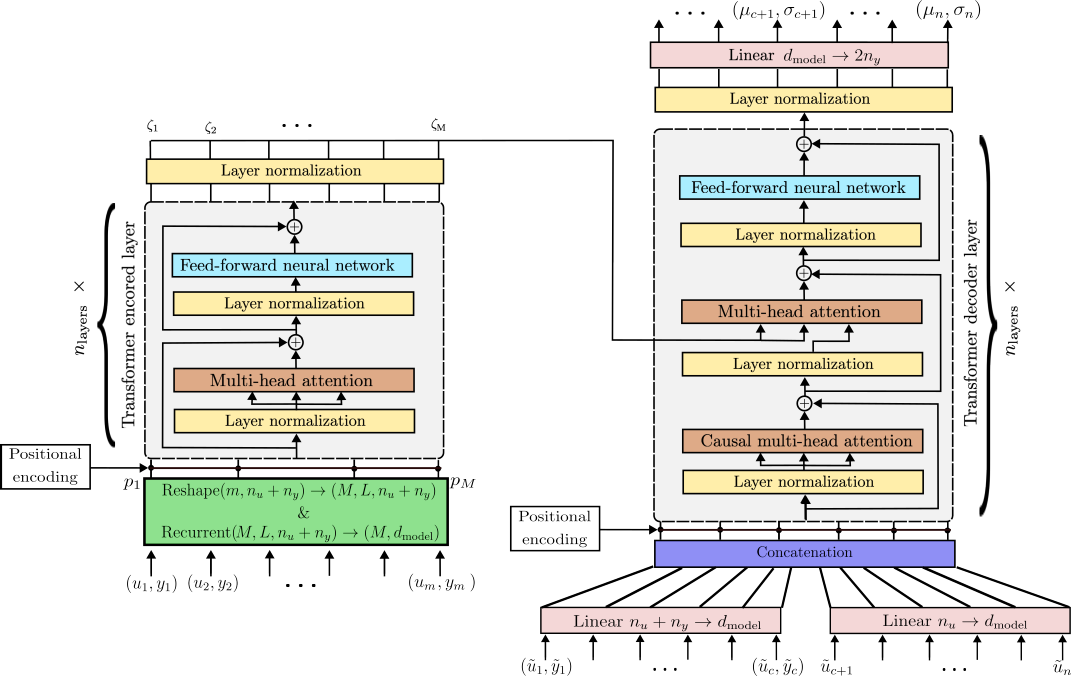}
\caption{Encoder-decoder Transformer for in-context system identification. The main differences with respect to the original Transformer architecture~\cite{vaswani2017attention} are highlighted in blue and pink. Additionally, an extra layer preceding the encoder backbone, referred to as recurrent patching neural network and highlighted in green, is incorporated to manage long context sequences.}
\label{fig:encoder_decoder_patch}
\end{figure*}

The meta model is implemented using an encoder-decoder Transformer architecture similar to the one introduced in~\cite{vaswani2017attention} for language translation, see Fig.~\ref{fig:encoder_decoder_patch} for a visual representation. 
With respect to~\cite{vaswani2017attention}, a recurrent patching network, described in details in Section~\ref{subsec:patching}, transforms the context data $(u_{1:m}, y_{1:m})$ into a shorter sequence $p_{1:M}$, with $p_i \in \mathbb{R}^{d_{\rm model}}$ and $M = \frac{m}{L}$, where  $L$ is an integer greater than $1$.
The encoder backbone then processes $p_{1:M}$ and maps it into $\zeta_{1:M}$, with  $\zeta_i \in \mathbb{R}^{d_{\rm model}}$(left side of Fig.~\ref{fig:encoder_decoder_patch}). 
Then, the decoder processes the query input $\tilde{u}_{\nin+1:\N}$ causally in time through masked self-attention, to generate the estimated distribution parameter $\mu_{\nin+1:\N},\sigma_{\nin+1:\N}$, incorporating the knowledge of $\zeta_{1:M}$ through a cross-attention mechanism.  Conceptually, the encoder can be interpreted as a system identification module, where $\zeta_{1:M}$ is an implicit representation of the system, and the decoder is a simulator.

Before the decoder block (right side of Fig.~\ref{fig:encoder_decoder_patch}), the initial conditions (namely, initial input/output measurements) are processed by a linear layer that maps each time step from $\mathbb{R}^{n_u + n_y}$ to $\mathbb{R}^{d_{\rm model}}$. Similarly, the query input $\tilde{u}_{\nin+1:n}$ undergoes a separate transformation from $\mathbb{R}^{n_u}$ to $\mathbb{R}^{d_{\rm model}}$ (bottom-right blocks in Fig.~\ref{fig:encoder_decoder_patch}). These processed sequences are concatenated to form a unique sequence of length $\N$, with each element having dimension $d_{\rm model}$. Then, a positional encoding is performed to preserve the temporal structure before the sequence enter the decoder backbone of the Transformer. The final layer of the decoder (upper-right block in Fig.~\ref{fig:encoder_decoder_patch}) outputs both the mean $\mu_{\nin+1:\N}$ and standard deviation $\sigma_{\nin+1:\N}$ of the probability distribution of the system output at each future time step $\nin+1,\ldots,\N$. 

Training of $\free_{\phi}$ is performed via stochastic (or mini-batch) gradient-descent, minimizing the  loss $J(\phi)$ in \eqref{eq:J_prob}. At each iteration, a batch of $b$ systems ${S}^{(i)}$ (with $i=1,\ldots,b$) is drawn from the class $p(S)$ and the corresponding dataset $\D^{(i)}$ is generated to compute the loss. 

\section{Handling long sequences} \label{sec:long_seq}
An intrinsic  limitation of the Transformer architecture is that the attention mechanism  has a computational complexity that grows quadratically with the sequence length.
This results in a limitation on the length of context sequences that the meta model $\free_{\phi}$ can efficiently handle. In this section, we propose two different approaches to address these limitations: one is applied during training, and the other during inference.

\subsection{Patching}
\label{subsec:patching}
\begin{figure}[!bt]
\centering
\includegraphics[width=.48\textwidth]{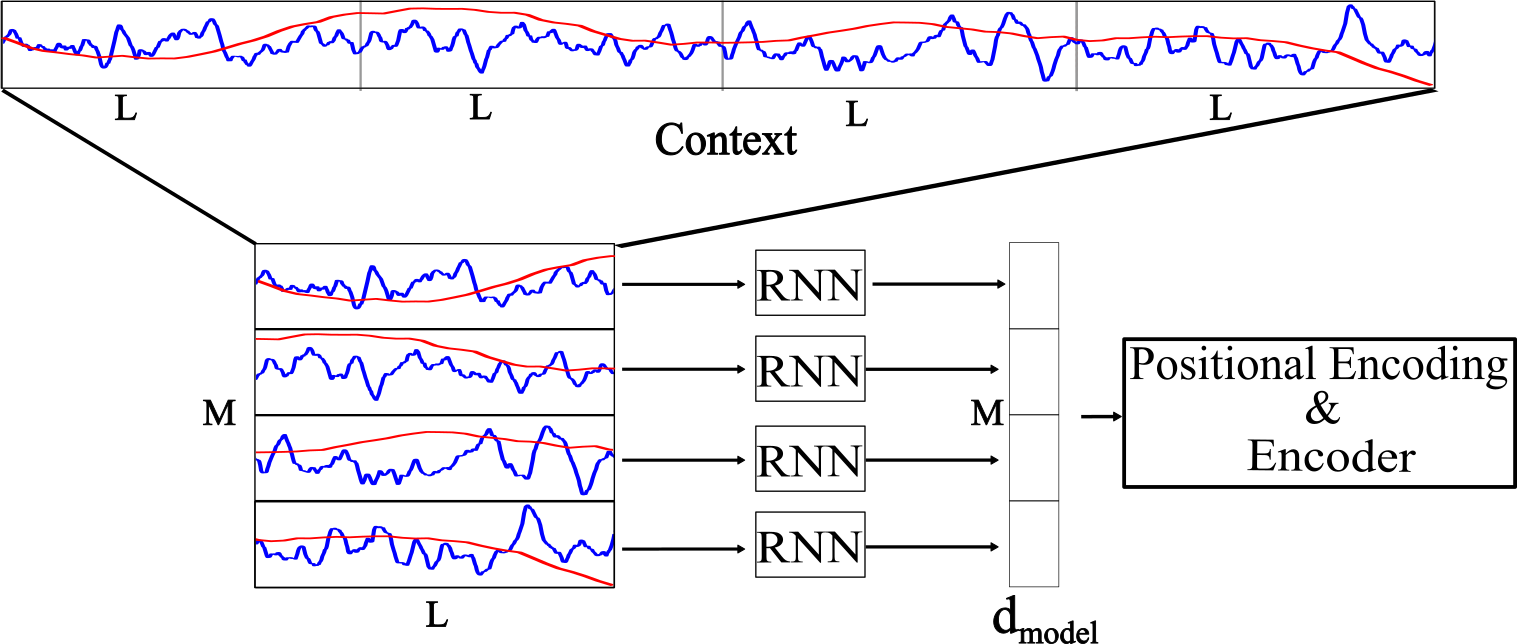}
\caption{Visual representation of the implemented patching approach. For clarity, a single-input (\(n_u=1\), blue) single-output (\(n_y=1\), red) system is depicted}
\label{fig:patch_structure}
\end{figure}

The patching approach, originally introduced in~\cite{dosovitskiy2021imagepatching} for image processing, was later adapted for time series data in~\cite{Nie2023timepatching}. This method partitions the context sequence into smaller sub-sequences, referred to as \emph{patches}, which are individually processed by a \emph{patching network} to reduce the time dimensionality. This approach significantly reduces the computational cost of training by shortening the sequence length before processing by the Transformer.  Specifically, the input/output context sequence \((u_{1:m}, y_{1:m})\) is divided into \(M\) non-overlapping patches,\footnote{For simplicity, we assume \(m\) is a multiple of \(M\).} each of length \(L = \frac{m}{M}\) and dimension \(n_u + n_y\). These patches are then processed by a recurrent neural network (RNN), which maps each patch to a single vector of dimension \(\dmodel\). The resulting \emph{patch embedding} sequence, \( p_{1:M} \), is enriched with positional encoding to take into account temporal order information before being passed to the encoder backbone (bottom-left green block in Fig.~\ref{fig:encoder_decoder_patch}). A visual representation of this approach is shown in Fig.~\ref{fig:patch_structure}.  

This approach reduces the sequence length processed by the multi-head attention mechanism by a factor of \(L\), leading to a computational complexity reduction of a factor $L^2$ for the training of the meta model \(\free_{\phi}\).  

\subsection{Ensembling}
\label{subsec:ensembling}

{Differently from the patching strategy, the ensembling method described in this section is applied only during the inference phase, i.e., after meta model training. 
The approach is based on averaging multiple predictions to improve robustness. 
This idea builds on the well-established principle that predictive performance can often be improved through ensembling, which combines the outputs of multiple models rather than relying on a single one~\cite[Chapter 14]{Bishop06ML}.}

{We consider the case where the Transformer has been trained on context sequences of length $m$, but a longer sequence $(u_{1:m'}, y_{1:m'})$ with $m' > m$ is available from the real system at inference time. 
From this longer sequence, we extract shorter overlapping subsequences $\{(u_{1:m}^i, y_{1:m}^i)\}_{i=1}^{s}$ of length $m$. 
}Considering all  overlapping sub-sequences, $s=m' - m + 1$. Each subsequence  $(u_{1:m}^i, y_{1:m}^i)$ is provided as context to the pre-trained Transformer, along with a fixed query input, to generate multiple predictions for the same query output, i.e., 
\begin{multline}   
    \label{eq:ensembles}
    \mu_{\nin+1:\N}^i, \sigma_{\nin+1:\N}^i =  \\ 
    \free_\phi(\tilde{u}_{\nin+1:\N}, \tilde{u}_{1:\nin}, \tilde{y}_{1:\nin}, u_{1:m}^i, y_{1:m}^i),\, i=1,\dots,s, 
\end{multline}
where $\mu_{\nin+1:\N}^i$ and $\sigma_{\nin+1:\N}^i$ represent the predicted mean and uncertainty for the query output based on the \(i\)-th context subsequence \((u_{1:m}^i, y_{1:m}^i)\).

{A single  prediction is then obtained by combining the $s$ predictions through a weighted average. Specifically, at a generic query time step $k$, the ensembled prediction is computed as:}
\begin{equation}
\label{eq:ensemble}
\mu_k = \sum_{i=1}^{s} \omega_{k}^i \mu_{k}^i.
\end{equation}
{where $\omega_{k}^i$ are non-negative weights that determine the contribution of each prediction. 
The goal is to select the weights to obtain an unbiased estimate of $\tilde{y}_k$ with minimum variance.}

From \eqref{prob_output_sample}, we may interpret $\mu_{k}^i$ as a ``noisy'' estimate of the true unknown output $\tilde y_k$:
\begin{equation}
\label{likelihood_mu}
    p(\mu_{k}^i) = 
    \mathcal{N}(\mu^i_k;\, \tilde y_k, (\sigma^i_k)^2).
\end{equation}
{Consequently, the ensembled mean $\mu_k$ in \eqref{eq:ensemble} is also Gaussian, with mean 
$\bar{\mu}_k = \tilde{y}_k \sum_{i=1}^{s} \omega_{k}^i$. 
To ensure that $\mu_k$ is an unbiased estimator of $\tilde{y}_k$, the weights must satisfy the constraint:}
\begin{equation}
\label{eq:constraint_sum}
\sum_{i=1}^{s} \omega_{k}^i = 1.
\end{equation}
Furthermore, assuming that $\mu_{k}^i$ are independent of each others for all $i=1,\dots,s$, the variance $\sigma_{\mu_k}^2$ of $\mu_k$ can be written:
\begin{equation}
    \label{eq:variance}
    \sigma_{\mu_k}^2 = \sum_{i=1}^s (\omega_{k}^i \sigma_{k}^i)^2. 
\end{equation}
Under constraint \eqref{eq:constraint_sum}, it can be proven that the variance $\sigma_{\mu_k}^2$ in \eqref{eq:variance} is minimized by the weights:
\begin{equation} \label{eqn:derweight}
    \omega_{k}^i = \frac{(\sigma_{k}^i)^{-2}}{\sum_{j=1}^s (\sigma_{k}^j)^{-2}}.
\end{equation}
A detailed derivation of \eqref{eqn:derweight} is reported in the appendix. 

Note that, in the case of $\sigma^i_{k}$ equal to each other for all $i=1,\dots,s$, then $\omega_{k}^i=\frac{1}{s}$ and $\mu_k$ in \eqref{eq:ensemble} is given by an unweighted  ensemble average.

\section{Synthetic data generation for system identification}
\label{sec:synthetic_data_generation}

In this section, we discuss how the pre-trained meta model can be leveraged for classical system identification, where the goal is to estimate a (parsimonious) model $G$ of a dynamical system $S_0$ based on observed data.

The use of synthetic data in this context is motivated by regimes in which the available real training dataset $\Dtr$ is small or weakly informative (e.g., limited excitation bandwidth or short experiments). In such cases, flexible parametric models $G$ trained only on $\Dtr$ tend to overfit and extrapolate poorly. The pre-trained meta model provides a class-level prior over plausible input–output trajectories, which we exploit to generate synthetic data conditioned on $\Dtr$ that enrich the training set and act as a regularizer for $G$.

This strategy rests on one key assumption: $S_o$ can be reasonably well described
by $\mathcal{M}_\phi$, i.e., $\mathcal{D}_{tr}$, used as context, allows the meta
model to meaningfully predict the trajectories of $S_o$.

In this setting, the meta model does not replace classical system identification but complements it. By generating synthetic trajectories conditioned on the same training data $\Dtr$, it creates additional scenarios tailored to the specific system of interest, rather than generic WH realizations. The resulting two-step pipeline, “generate context-aware synthetic data $\rightarrow$ estimate a parametric model”, therefore trains $G$ on a richer dataset composed of $\Dtr$ together with synthetic trajectories adapted to the target system, instead of training $G$ directly on $\Dtr$ combined with unconditioned WH simulations.

Let us consider an observed input/output data sequence $(u_{1:m}, y_{1:m})$ (referred so far as context sequence) generated by system of interest $\So$. Since this dataset represents the available information about $\So$, it can be used either as a context for the meta model or as a training dataset for estimating a model $\G$ of the system $\So$, or both. For this reason, $(u_{1:m}, y_{1:m})$ will be interchangeably referred to as the ``context dataset'' or ``training dataset'', and will be denoted as $\Dtr$.

We consider a parametric model $\G$ characterized  by a parameter vector $\theta$, where $\hat{y}_t(\theta)$ represents the output of $\G$ for a given input sequence up to time $t$, i.e.,
\begin{equation}
    \hat{y}_t(\theta) = \G(u_{1:t}; \theta).
\end{equation}
For simplicity of exposition, we assume a single-output system. 

According to classical parametric system identification, the model parameters $\theta$ are typically obtained by minimizing an average loss function $\ell$ over the training dataset $\Dtr$, i.e., 
\begin{equation} \label{eq:par:id}
    \hat{\theta} = \arg \min_{\theta}  \frac{1}{m} \sum_{t=1}^m \ell(y_t, \hat{y}_t(\theta)). 
\end{equation}

To enrich the training set \( \Dtr \), we aim to generate an additional, potentially  infinite, set of \emph{synthetic} input/output sequences $ \tilde{u}^{i}_{1:\tilde{T}},  \tilde{y}^{i}_{1:\tilde{T}}$ where \( i = 1, 2, \ldots \) is used to represent the $i$-th synthetic trajectory and \( \tilde{T} \) is the length of the synthetic trajectories, that for simplicity of notation, is assumed to be the same for all synthetic datasets. The synthetic trajectories are drawn from a  common probability distribution \( p(\tilde{u}_{1:\tilde{T}}, \tilde{y}_{1:\tilde{T}}) \), ideally matching the one underlying \( \Dtr \).

The main idea behind the data augmentation strategy proposed in the paper is  to use the pre-trained meta model $\free_{\phi}$ to  generate the synthetic input/output trajectories. Indeed, the meta model has  the ability to infer the behavior of the specific query system $\So$ directly from the training (or, equivalently, context) dataset \( \Dtr \), which implicitly encodes key characteristics of the system \( \So \). This trained meta model can then predict the output \( \tilde{y}_{1:\tilde{T}} \) for any synthetic input sequence \( \tilde{u}_{1:\tilde{T}} \), thereby facilitating generation of a potentially infinite dimensional set of synthetic data, by implicitly transferring knowledge across systems within the same class.

The following sections discuss two approaches for combining the training dataset $\Dtr$ with synthetic data generated by the pre-trained meta model $\free_\phi$. These approaches are related, as both incorporate the synthetic dataset as an additional regularization term in the original loss function in \eqref{eq:par:id}. However, they differ in how this regularization term is constructed. In the first approach, discussed in Section~\ref{sec:syn_determin} and originally introduced in \cite{piga2024synthetic},  equal weights in the regularization term are assigned  to all synthetic samples. In contrast, the second approach, presented in Section~\ref{sec:syn_ML}, derives the regularization term based on a maximum-likelihood formulation and considers the uncertainty of the synthetic samples in constructing the training loss. 

\subsection{Deterministic regularization} \label{sec:syn_determin}

The parameters $\theta$ that define the model $\G(\cdot,\theta)$ of the system $\So$ are estimated  by minimizing an expected loss that incorporates both the real data $\Dtr$ generated by the system \( \So \) and synthetic data generated by the pre-trained meta model $\free_{\phi}$, i.e., 
\begin{align}
    \hat{\theta} = \arg \min_{\theta} & \frac{1}{m} \sum_{t=1}^m \ell(y_t, \hat{y}_t(\theta)) + \nonumber \\
    & + \lambda \, \mathbb{E}_{p(\tilde{u}_{1:\tilde{T}}, \tilde{y}_{1:\tilde{T}})} \left[ \frac{1}{\tilde{T}} \sum_{k=1}^{\tilde{T}} \ell(\tilde{y}_k, \hat{\tilde{y}}_k(\theta)) \right], \label{eqn:lossE}
\end{align}
where \( \hat{\tilde{y}}_k(\theta) \) is the model’s estimation output for a synthetic input sequence \( \tilde{u}_{1:k} \), i.e., $\hat{\tilde{y}}_k(\theta)=\G(\tilde{u}_{1:k},\theta)$.

The regularization hyper-parameter \( \lambda \geq 0 \) controls the relative importance of real versus synthetic data. Setting \( \lambda = 0 \) results in training on real data only, which may cause overfitting, particularly with complex models or limited datasets. A higher \( \lambda \) value emphasizes synthetic data, which can introduce distributional biases from the synthetic data generation process. Thus, \( \lambda \) should be chosen based on the quantity of real training data and on the quality of the synthetic data. 
In this work, \( \lambda \) is chosen via hold-out validation. Early stopping over a validation set is used to prevent overfitting, which may occur  in case of small-size \( \Dtr \) and when synthetic data are not employed. To not use  additional data beyond what is already used for model estimation with early stopping, the same validation set can also be used for tuning $\lambda$.

The expected value in \eqref{eqn:lossE} is approximated with a sample average, by applying \( \bsy \) synthetic input sequences \( \tilde{u}^{i}_{1:\tilde{T}} \) (for \( i = 1, \ldots, \bsy \)) as input in the decoder, generating the corresponding outputs \( \tilde{y}^{i}_{1:\tilde{T}} \)\footnote{As initial conditions in the generation of the output trajectory $\tilde{y}^{i}_{1:\tilde{T}}$ we can consider  the last $c$ samples of the context  $\Dtr$.}. Consequently, the estimation problem becomes:
\begin{align}
    \hat{\theta} = \arg \min_{\theta} & \frac{1}{m} \sum_{t=1}^m \ell(y_t, \hat{y}_t(\theta)) + \nonumber \\
    & + \lambda \frac{1}{\bsy} \frac{1}{\tilde{T}} \sum_{i=1}^{\bsy} \sum_{k=1}^{\tilde{T}} \ell(\tilde{y}^{i}_k, \hat{\tilde{y}}^{i}_k(\theta)). \label{eqn:loss_sample}
\end{align}

The optimization problem \eqref{eqn:loss_sample} can solved using mini-batch stochastic gradient descent, with \( \bsy \) new synthetic input/output sequences generated at each iteration.

\subsection{Maximum-likelihood based regularization} \label{sec:syn_ML}

In this section, we present an alternative approach based on maximum likelihood for integrating  real training data $\D_{\text{tr}}$ and synthetic data   into a unified loss function for estimating the parameters $\theta$ of the model $\G$. 

We  assume that  there exists an unknown ``true'' parameter $\theta^o$ such that, for any given input sequence $u_{1:t}$, the true system output $y_t^{\rm o}$ at time $t$ is given by:
\begin{align}
    y_t^{\rm o} = \G(u_{1:t},\theta^o).
\end{align}

The measured training output is then modeled as:
\begin{align}
    y_t = \underset{y_t^{\rm o}}{\underbrace{\G(u_{1:t},\theta^o)}} + e_t, \quad t=1,\ldots,m,
\end{align}
where $e_t$ is \emph{independent and identically distributed} (i.i.d.) Gaussian noise, i.e.,  $ e_t \sim \mathcal{N}(0, \sigma^2_e)$.

Thus, the likelihood function of the model parameters $\theta$ given the training dataset $\D_{\text{tr}}$ can be factorized as:
\begin{align} \label{eqn:ysynt_mu}
    p(y_{1:m}|u_{1:m},\theta) = \prod_{t=1}^m \mathcal{N}(y_t; \G(u_{1:t},\theta), \sigma^2_e).
\end{align}
The predicted mean provided by the meta model is used as synthetic output $\tilde{y}_{k}$, namely:
\begin{align}\label{eqn:ysyn}
    \tilde{y}_{k}(\tilde{u}_{1:k},\D_{\text{tr}}) = \mu_k(\tilde{u}_{1:k},\D_{\text{tr}}), \ \ k=1,\ldots,\tilde{T},
\end{align}
where we emphasize the dependence of $\mu_k$ (and consequently $\tilde{y}_{k}$) on the training dataset $\D_{\text{tr}}$ (or, equivalently, the context). 

According to the meta modeling assumption in \eqref{eq:fact_q}, the true output $\tilde{y}_k^{\rm o}=\G(\tilde{u}_{1:k},\theta^o)$ of the system of interest $\mathcal{S}_o$ for a query input $\tilde{u}_{1:k}$ follows a Gaussian distribution:
\begin{align}
    p(\tilde{y}_k^{\rm o}|\tilde{u}_{1:k},\D_{\text{tr}}) = \mathcal{N}(\tilde{y}_k^{\rm o}; \mu_k, \sigma^2_k),
\end{align}
and
\begin{align}
    p(\tilde{y}_{1:\tilde{T}}^{\rm o}|\tilde{u}_{1:\tilde{T}},\D_{\text{tr}}) = \prod_{k=1}^{\tilde{T}} \mathcal{N}(\tilde{y}_k^{\rm o}; \mu_k, \sigma^2_k).
\end{align}
In the above equations, the dependence of $\mu_k$ and $\sigma^2_k$ is omitted not to burden the notation.

From the choice of the synthetic output $\tilde{y}_k$ in \eqref{eqn:ysyn}, and due to the symmetry of the Gaussian distribution, the likelihood function of the model parameters $\theta$ given the single synthetic output sequence $\tilde{y}_{1:k}$ and the training dataset $\Dtr$ is:
\begin{align}
    & p(\tilde{y}_{1:\tilde{T}}|\tilde{u}_{1:\tilde{T}},\D_{\text{tr}},\theta) = \nonumber \\
    & \quad \quad \prod_{k=1}^{\tilde{T}} \frac{1}{\sqrt{2\pi \sigma_k^2}} \exp\left(-\frac{(\mu_k - \G(\tilde{u}_{1:k},\theta))^2}{2\sigma_k^2}\right).
\end{align}

Thus, the joint likelihood function of the model parameters $\theta$ given both the training and synthetic output is:
\begin{align}
    & p(y_{1:m}, \tilde{y}_{1:\tilde{T}}| u_{1:m}, \tilde{u}_{1:\tilde{T}},\theta) = \nonumber \\
    & p(y_{1:m}| u_{1:m},\theta) p(\tilde{y}_{1:\tilde{T}}| \tilde{u}_{1:\tilde{T}}, \underset{\D_{\text{tr}}}{\underbrace{u_{1:m},y_{1:m}}}, \theta) = \nonumber \\
    & \prod_{t=1}^m \mathcal{N}(y_t; \G(u_{1:t},\theta), \sigma^2_e) 
    \prod_{k=1}^{\tilde{T}} \mathcal{N}(\tilde{y}_{k};  \G({\tilde{u}}_{1:k},\theta), \sigma_{k}^2) = \nonumber \\
    & \prod_{t=1}^{m} \frac{1}{\sqrt{2\pi \sigma_e^2 }} \exp\left(-\frac{(y_t -  \G(u_{1:t},\theta)  )^2}{2\sigma_e^2}\right) \times \nonumber \\
    & \prod_{k=1}^{\tilde{T}} \frac{1}{\sqrt{2\pi \sigma_k^2}} \exp\left(-\frac{(\mu_k - \G(\tilde{u}_{1:k},\theta))^2}{2\sigma_k^2}\right).
\end{align}

Thus, the log-likelihood function is given by:
\begin{align}
    & \log p(y_{1:m}, \tilde{y}_{1:\tilde{T}}| u_{1:m}, \tilde{u}_{1:\tilde{T}},\theta) = \nonumber \\
    & - \sum_{t=1}^{m} \frac{(y_t -  \G(u_{1:t},\theta))^2}{2\sigma_e^2}   
    - \sum_{k=1}^{\tilde{T}} \frac{(\mu_k - \G(\tilde{u}_{1:k},\theta))^2}{2\sigma_k^2} + C,
\end{align}
where $C$ is a constant term that does not depend on the model parameters $\theta$.

Maximizing the log-likelihood with respect to $\theta$ is equivalent to solving the following optimization problem:
\begin{align} 
    \min_{\theta} \sum_{t=1}^{m} \frac{(y_t -  \G(u_{1:t},\theta))^2}{2\sigma_e^2}  
    + \sum_{k=1}^{\tilde{T}} \frac{(\mu_k - \G(\tilde{u}_{1:k},\theta))^2}{2\sigma_k^2}. \label{eqn:syntML}
\end{align}
Since the noise variance $\sigma_e^2$ is constant (and might also be  unknown),  problem  \eqref{eqn:syntML} is equivalently rewritten by introducing a hyper-parameter $\lambda$ that weighs the second summation, i.e.,
\begin{align}
    \min_{\theta} \sum_{t=1}^{m} (y_t -  \G(u_{1:t},\theta))^2  
    + \lambda \sum_{k=1}^{\tilde{T}} \frac{(\mu_k - \G(\tilde{u}_{1:k},\theta))^2}{\sigma_k^2},
\end{align}
with $\lambda$ tuned through cross-validation to balance the contribution of the synthetic data term, as also discussed in Section~\ref{sec:syn_determin}. 

The extension to the case of multiple synthetic sequences is straightforward, leading to:
\begin{align} \label{eqn:optML}
    \min_{\theta}  \sum_{t=1}^{m} (y_t -  \G(u_{1:t},\theta))^2 
    + \lambda \sum_{i=1}^{q} \sum_{k=1}^{\tilde{T}} \frac{(\mu_k^i - \G(\tilde{u}_{1:k}^i,\theta))^2}{(\sigma_k^{i})^2}.
\end{align}
Summarizing, the optimization problem \eqref{eqn:optML} aims at minimizing the sum of squared errors, where the first term accounts for the discrepancy between the model predictions and the observed real data, while the second term incorporates synthetic data. The synthetic data contributions are weighted inversely to their associated uncertainty, as reflected by the variance terms $(\sigma_k^{i})^2$. Consequently, synthetic samples with high uncertainty have a lower influence on the loss, ensuring that the estimation process prioritizes more reliable data points. 

The effectiveness of synthetic data is therefore not assumed a priori, but controlled at the point of use through this uncertainty weighting. The probabilistic meta model provides a per-sample uncertainty~$\sigma_k$, and the maximum-likelihood loss weights synthetic contributions inversely to~$\sigma_k^2$, so that synthetic samples deemed uncertain have limited impact on the estimation of $G$, while high-confidence synthetic predictions carry more weight. In addition, before relying heavily on synthetic trajectories, the practitioner can perform simple consistency checks on their distribution (e.g., input spectrum, amplitude range, operating region) relative to the available real data, and restrict the use of synthetic data to regimes where the meta model is both confident and the synthetic inputs remain within the domain explored by the real system.

\section{Examples} \label{sec:results}

In this section, we demonstrate the capabilities of the  meta modelling paradigm through several simulation examples, as well as system identification benchmarks using real data.

Trainings and inferences are performed using an Nvidia RTX 4090 GPU with 24GB of RAM. To ensure the reproducibility of all experiments, the complete codebase, including dataset generation, training routines, and trained meta models, has been made publicly accessible in a GitHub repository~\cite{repo}.

\subsection{Meta model training}

\subsubsection{System class}
The meta model $\free_\phi$ is trained on synthetic datasets generated by simulating Wiener-Hammerstein (WH) systems characterized by the structure $G_1\!-\!F\!-\!G_2$, where $F$ represents a static nonlinearity positioned between two LTI blocks, $G_1$ and $G_2$. The discrete-time LTI blocks $G_1$ and $G_2$ are randomly generated, with their orders uniformly sampled in the range $[1,10]$, ensuring variability in their dynamic behavior. The pole locations of $G_1$ and $G_2$ are parameterized by magnitudes and phases, which are independently drawn from uniform distributions over the intervals $(0.5, 0.97)$ and $(0, \pi/2)$, respectively. The nonlinear function $F$ is implemented as a feedforward neural network with a single hidden layer that contains $32$ hidden units. The network weights are drawn from a Gaussian distribution with Kaiming scaling~\cite{he2015delving}.

This choice of training class is principled rather than arbitrary. 
Any fading-memory nonlinear system admits an arbitrarily close approximation 
by a finite-order Volterra series~\cite{boyd1985fading}, and parallel 
Wiener–Hammerstein structures can exactly represent finite-order Volterra 
kernels~\cite{billings1982identification}. 
As a result, the WH-based synthetic corpus spans a rich family of nonlinear 
dynamics and provides a domain-informed pre-training distribution, rather than 
a system-specific simulator.

The inputs $u_{1:m}$, used as context signals, are discrete multisine signals of length $m$, with unitary amplitude and random phase, resulting in a flat spectral density. The active frequency components are constrained within a passband defined by the spectral index range $[f_{\text{min}}, f_{\text{max}}]$, with $f_{\text{min}} \geq 1$ and $f_{\text{max}} \leq \lceil m/2 \rceil$,  with $\lceil \cdot \rceil$ denoting the ceiling operator, and $\lceil m/2 \rceil$ corresponding to the Nyquist frequency for a discrete signal. The upper bound $f_{\text{max}}$ is randomly selected from a uniform distribution over the range $[1, \lceil m/2 \rceil]$. The lower bound $f_{\text{min}}$ is then assigned based on the following scheme:

\begin{itemize}
    \item with probability $0.5$, we set $f_{\text{min}} = 1$, enforcing a low-pass filtering effect;
    \item with probability $0.5$, we sample $f_{\text{min}}$ from a uniform distribution over $[1, f_{\text{max}}]$.
\end{itemize}

The signals are synthesized in the frequency domain and transformed into the time domain using an inverse Discrete Fourier Transform. Finally, they are normalized to maintain zero mean and unit standard deviation. Query input signals $\tilde{u}_{1:\N}$ are constructed similarly.

To simulate measurement errors coming from sensors, white Gaussian noise with a standard deviation of $\sigma_{\text{noise}} = 0.01$ is added to the outputs.

In the experiments presented here, the WH-based synthetic corpus is generated by a parametric model. In practice, the same meta-learning framework can be instantiated using any available simulator or physics-based model of a target application domain (e.g., robotics, powertrain, process control). The essential requirement is access to a simulator that captures a class of plausible systems, not necessarily a perfect representation of the specific plant of interest. As we will discuss in Section~\ref{sec:model_adap}, discrepancies between the simulator and the real system can be compensated by lightweight adaptation using a relatively small amount of data.

\begin{remark}
In contrast to large-scale foundation models trained on heterogeneous data with billions of parameters, the proposed meta model has a modest parameter count and is trained on synthetic trajectories drawn from a physically motivated class of systems. The WH-based design leverages existing knowledge on universal approximation for fading-memory nonlinear systems, rather than relying solely on brute-force scale.
\end{remark}
\subsubsection{Meta model architecture}
The  Transformer architecture representing the meta model $\free_\phi$ comprises $n_{\rm layers} = 12$ layers, each with $\dmodel = 128$ hidden units. Attention mechanisms are implemented with $n_{\rm heads} = 4$ in both the encoder and decoder. The query input/output sequences consists $\N = 130$ samples, with the initial ${\nin} = 30$ samples used as initial conditions.

To assess performance across different context sequence complexities, two meta models are trained: one for a context length of $m = 400$ and another for $m = 16 000$. This setup enables exploration of how the length of the context sequence affects model accuracy and computational efficiency. For the extended context length of $m = 16000$, the Transformer architecture incorporates the recurrent patching approach described in Section~\ref{subsec:patching}. The context is divided into $M = 400$ patches, each of length $L = m / M = 40$. These patches are sequentially processed by a vanilla RNN with a single hidden layer with $\dmodel = 128$ units. The final hidden state of each patch is further transformed through a square linear layer of size $\dmodel$, producing a sequence $p_{1:M}$. This sequence is then used as input to the Transformer's encoder backbone. Conversely, for the shorter context length of $m = 400$, the recurrent patching method is not employed. Instead, the encoder input sequence $p_{1:M}$, where $M = m$, is directly obtained by processing the input/output samples, $u_{1:m}$ and $y_{1:m}$, through a linear layer that maps $n_u + n_y$ inputs to $\dmodel$ outputs. In both scenarios, the Transformer architecture is described by approximately $5.5$ million parameters, with  additional $16896$ parameters for the RNN layer used in the patching strategy.

The meta model $\free_\phi$ is trained to minimize the negative log-likelihood loss $J_\phi$  defined in \eqref{eq:J_prob} across $3$ million epochs, utilizing the AdamW optimizer~\cite{loshchilov2018adamW} with minibatches of size $b = 32$. The training times for the meta models with context lengths of $m = 400$ and $m = 16000$ are $28$ and $50$ hours, respectively.

\subsection{Meta model testing}

The predictive performance  of the pre-trained meta model $\free_\phi$ is assessed  not only across realizations  of WH systems belonging to the same class used for meta model training, 
but also on simulated systems from a different class excited with non multisine input signals, as well as on 
\textit{real} datasets  from the nonlinear system identification benchmarks~\cite{NonLinearBench}. 

For the meta model handling a long context length ($m = 16000$), the patching-based  approach is adopted, while  ensembling  is used when   the meta model handling a shorter context ($m=400$) is considered. This allows us to leverage the full context data, in case the available training dataset, at inference, is larger than $400$ samples.

\subsubsection{In-distribution test data}

First, the performance of the pre-trained meta model is evaluated on test samples drawn from the same distribution as the training data. To this aim,  $320$ random realizations of simulated WH systems are generated for testing.

Table~\ref{tab:example_adhoc} reports the mean and standard deviation of the Root Mean Squared Error (RMSE)  achieved over the 320 realizations, along with average inference times to predict the output on a query window of length $\N-\nin = 100$. The patching and ensembling approaches achieve similar predictive performance, with the patching approach being approximately 2 times faster in making inferences.

\begin{figure}[!bt]
\centering
\includegraphics[width=.48\textwidth]{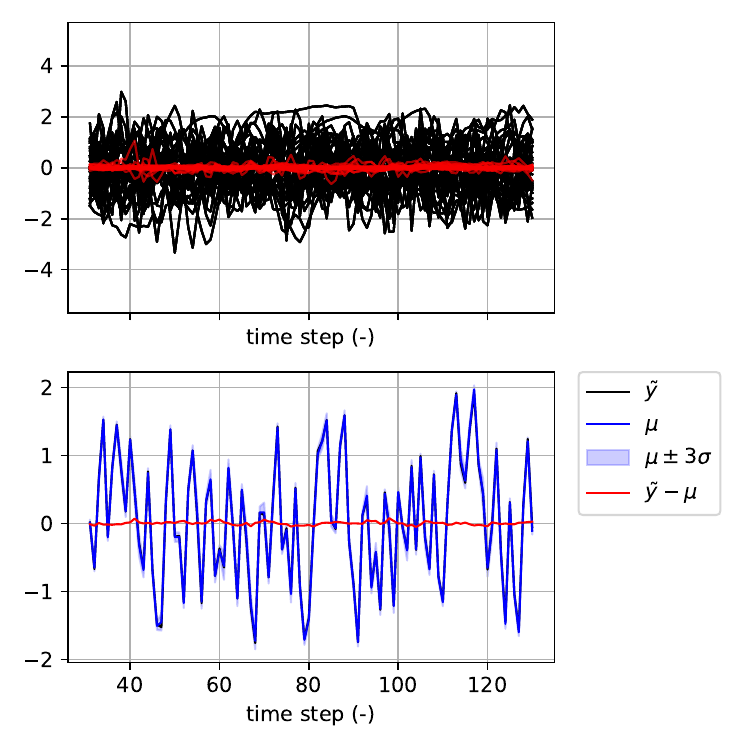}
\caption{{In-distribution test results obtained using the patching-based meta model.
\textit{Top:} aggregated prediction results over 320  WH systems, highlighting the consistency and stability of the predictions across many system realizations. 
\textit{Bottom:} representative WH system showing the true output $y$, the predicted mean $\mu$, the prediction error $y - \mu$, and the uncertainty band $\mu \pm 3\sigma$. 
The narrow confidence intervals and low error illustrate accurate prediction.}}

\label{fig:WH_in_distr}
\end{figure}

Figure~\ref{fig:WH_in_distr} illustrates the performance of the trained meta model. For visualization purposes, only results from the patching approach are displayed. {The top panel shows the aggregated simulation results over the 320 randomly sampled Wiener–Hammerstein systems from the same distribution used for meta model training, highlighting the consistency of the model predictions across systems. } The bottom panel provides a detailed view of a single WH realization, showing the true output $\tilde{y}_{\nin+1:\N}$, the predicted mean output $\mu_{\nin+1:\N}$,  the corresponding prediction error $\tilde{y}_{\nin+1:\N} - \mu_{\nin+1:\N}$, and the  uncertainty intervals computed as $\mu_{\nin+1:\N} \pm 3\sigma_{\nin+1:\N}$. {The tight uncertainty bounds  and low error in the figure demonstrate that the meta model accurately captures the system dynamics and provides well-calibrated uncertainty estimates}

\begin{table}[bt!]
    \centering
    \caption{In-distribution systems: Root Mean Square Error (RMSE) and inference time for patching-based and ensembling approach.}
    \begin{tabular}{ccc}
        \toprule
        Method & RMSE mean (std)   & Inference time [s] \\
        \midrule
        Patching & 0.048 (0.044) &  0.019 \\
        Ensembling & 0.038 (0.050) &   0.039 \\
        \bottomrule
    \end{tabular}
    \label{tab:example_adhoc}
\end{table}

\subsubsection{Benchmarks}

The pre-trained meta model $\free_\phi$ is further evaluated on publicly available nonlinear system identification benchmarks, namely the WH~\cite{schoukens2009wiener}, cascaded tanks with overflow~\cite{CascadedTanks}, and coupled electric drives (CED)~\cite{CED}, with CED  consisting of 2 input/output trajectories. 
The adopted benchmarks consist of \textit{real} datasets and thus allow us to investigate the meta model's ability to generalize beyond its training domain. Notably, these datasets are used solely for testing, with no additional training performed on the benchmark-specific training data. The available training datasets serve only as context for the pre-trained meta model $\free_\phi$. In our analysis, we prioritize demonstrating the zero-shot capabilities of the meta modeling paradigm, rather than directly comparing predictive performance with existing system identification methodologies.

The benchmark systems are chosen to span a spectrum of structural proximity to the synthetic training distribution, in order to stress-test the framework under varying degrees of distribution shift:
\begin{itemize}
    \item \textbf{Wiener–Hammerstein (WH) benchmark.}
    This is an \emph{in-distribution} test: the synthetic pre-training data are WH systems, so performance here validates that the framework correctly exploits the pre-training prior on systems drawn from the same class but generated in an experimental setup.
    \item \textbf{Coupled Electric Drives (CED).}
    The CED system is structurally close to a WH model: it consists of a linear dynamical core followed by a static nonlinearity, which matches the WH cascade structure.
    Good zero-shot performance on CED therefore confirms that the learned prior transfers to near-distribution systems with similar compositional structure.
    \item \textbf{Cascaded tanks with overflow.}
    This system exhibits a piecewise-linear overflow nonlinearity embedded in a nonlinear state-space model that does not factorize as a WH cascade, making it a genuinely out-of-distribution test for the meta model.
    The cascaded tanks benchmark is also characterized by a particularly small real training dataset, a regime where classical identification methods are prone to overfitting and struggle to generalize.
    In this setting, the inductive bias acquired during WH meta-training acts as a strong regularizer, enabling the framework to achieve competitive, and in some cases state-of-the-art, performance despite the distributional mismatch.
\end{itemize}

These observations highlight a regime where the proposed meta-modeling framework is particularly effective: when real training data per system are limited, the class-level prior learned from WH-based synthetic data provides a strong regularization effect and improves generalization. Conversely, on benchmarks with abundant task-specific data (e.g., systems with very long and informative training records), carefully tuned classical identification methods can still achieve higher asymptotic accuracy, whereas the meta model primarily offers the advantage of zero-shot prediction with modest performance gaps.

According to the benchmark guidelines, the first $50$, $5$, and $4$ samples of the test data can be used as initial conditions for assessing predictive performance on the WH, cascaded tank, and CED benchmarks, respectively. As outlined in the previous section, the meta model is trained to utilize $\nin = 30$ initial conditions for inference. For the WH benchmark where $50$ initial conditions are provided, only the last $30$ samples are used. Conversely, for the cascade tank benchmark, zero-padding is applied to construct the sequence of initial conditions fed to the meta model. For the CED benchmark, where the training and test sequences are known to be contiguous, the sequence of initial conditions is constructed by concatenating the final $26$ samples of the training data with the first $4$ samples of the test.

Since the meta model is able to simulate the output over a query window of size $100$, in order to accommodate the varying test sequence lengths in the benchmarks, an iterative inference approach is adopted. For the first inference step, the initial conditions provided by the benchmark dataset are used. In subsequent iterations, the $\nin$ final simulated output samples serve as initial conditions for the next test sub-sequence. This iterative  approach, combined with the weighted ensembling strategy, significantly increases the computational time required for inference. To address this limitation, the following strategies are implemented:

\begin{itemize}
\item \textbf{parallelization}: the predictions for the same inference window across all $400$-length context sub-sequences, used for the ensembling, are computed in parallel.
\item \textbf{pre-computation of encoder output}: during each inference iteration, only the decoder inputs change across different iterations. Consequently, the encoder output remains constant, allowing the encoder output to be computed only once for all the inference iterations.
\end{itemize}

\begin{figure}[!bt]
\centering
\includegraphics[width=.45\textwidth]{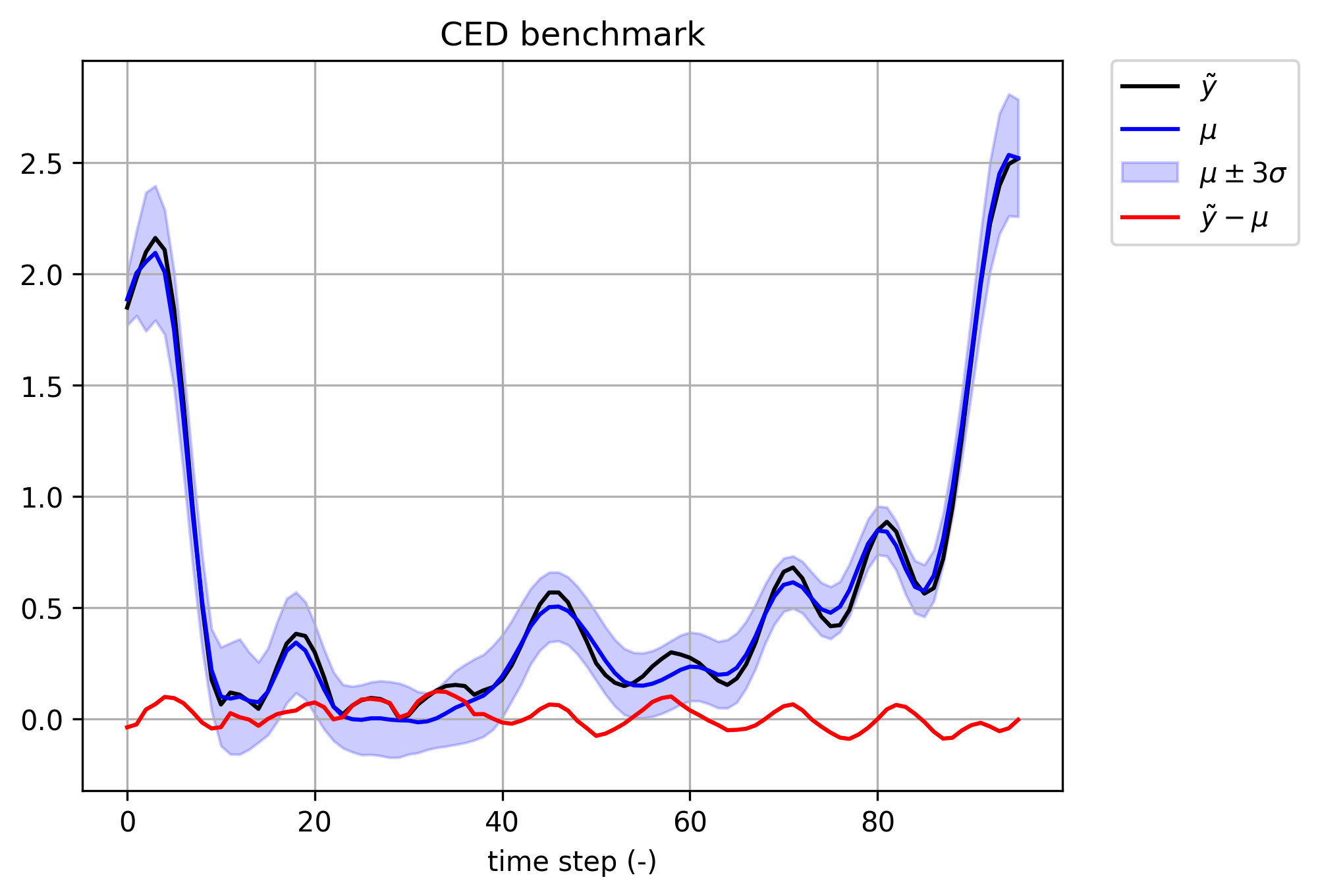}
\caption{{Prediction results on the CED benchmark for the second output signal. 
Comparison between true output $y$ (black) and meta model’s predicted mean $\mu$ (blue), along with simulation error $y - \mu$ (red) and uncertainty region $\mu \pm 3\sigma$ (shaded blue area). Most predictions lie within the confidence bounds, demonstrating robust zero-shot generalization to real data.}
}

\label{fig:CED_benchmark}
\end{figure}
\begin{table}[htb!]
\centering
\caption{Benchmarks: RMSEs and inference time over the test datasets for multiple methods.}
\small
\setlength{\tabcolsep}{3pt}
\begin{tabular}{cccc}
    \toprule
    Benchmark & Method & RMSE & \makecell{Training/Meta-\\inference time} \\
    \midrule
    \multirow{4}{*}{WH}
        & In-context  & 0.0014 & 5.91 s \\
        & dynoNet~\cite{forgione2021dynonet} & 0.0017 & 1-60 min \\
        & LSTM~\cite{champneys2024baseline} & 0.0015 & 1-24 hr \\
        & ARX~\cite{champneys2024baseline} & 0.0434 & 1-60 min \\
    \midrule
    \multirow{4}{*}{\makecell{Cascaded\\Tanks}}
        & In-context  & 0.429 & 0.22 s \\
        & dynoNet~\cite{forgione2021dynonet} & 0.421 & 1-60 min \\
        & LSTM~\cite{champneys2024baseline} & 0.452 & 1-24 hr \\
        & ARX~\cite{champneys2024baseline} & 0.697 & 1-60 min \\
    \midrule
    \multirow{4}{*}{CED}
        & In-context & [0.068, 0.055] & 0.17 s \\
        & dynoNet~\cite{forgione2021dynonet} & [0.062, 0.047] & 1-60 min \\
        & LSTM~\cite{champneys2024baseline} & [0.139, 0.111] & 1-24 hr \\
        & ARX~\cite{champneys2024baseline} & [0.433, 0.179] & 1-60 min \\
    \bottomrule
\end{tabular}
\label{tab:benchmarks}
\end{table}

{Table~\ref{tab:benchmarks} reports the RMSE values obtained on the test datasets for the three considered benchmarks, comparing the proposed approach with several baseline methods. 
It is important to note that the last column of the table compares the training time of the classical system identification baselines with the \emph{meta-inference} time of the proposed approach, rather than its meta-training time. 
As discussed in Section~\ref{sec:Framework}, the proposed meta model is trained exclusively on synthetic data and does not use benchmark data during training. 
Consequently, the only stage in which benchmark data are employed is during inference. 
For this reason, the comparison between baseline training time and meta-inference time provides a fairer assessment of computational efficiency, while the meta-training time, already reported in the paper, represents an offline cost incurred only once.}
It is worth noting that, although the main advantage of the proposed meta modeling paradigm is its ability to perform prediction in a zero-shot learning fashion, the predictive performance is also in line with state-of-the-art system identification algorithms (see leaderboard in~\cite{SchoukensTiels2024Results}).

Figure~\ref{fig:CED_benchmark} shows the second output trajectory of the CED benchmark, where we can see the true output $\tilde{y}_{\nin+1:n}$ of the test dataset, the predicted mean output $\mu_{\nin+1:n}$ generated by the meta model; the associated prediction error $\tilde{y}_{\nin+1:n} - \mu_{\nin+1:n}$; and the $99.7\%$ uncertainty interval computed as $\mu_{\nin+1:N} \pm 3 \sigma_{\nin+1:N}$. {
The simulation error $y - \mu$ (red) remains small and within the uncertainty band for most time steps, indicating reliable generalization to real-world data despite the model being trained solely on simulated systems. 
This result highlights the zero-shot learning capability of the pre-trained meta model.}

\subsubsection{Model adaptation and fine-tuning}
\label{sec:model_adap}

The performance of the pre-trained meta model is also evaluated in an out-of-distribution scenario characterized by shifts in both the input signals and the system class compared to those used for meta model training. To this end, we employed a simulator of a \emph{continuous stirred tank reactor} (CSTR) with jacket cooling, where a first-order irreversible reaction, $A \xrightarrow{} B$, occurs. The physical dynamics of the system are detailed  in~\cite[Chapter 3]{Luyben2007ChemCont}. In our analysis, the input variable is the flow rate of the cooling medium within the jackets, while the output is the concentration of reactant $A$ in the reactor. The output is normalized to have $0$ mean and unitary standard deviation.  The reactor temperature and the temperature of the cooling medium are considered as system states. 

The input to these systems is a \emph{random binary signal} (RBS), used to control the flow of the cooling medium inside the jackets. The context sequence length is set to $m = 16000$, leveraging the recurrent patching-based approach, while the number of initial conditions is set to $\nin = 30$. To emulate realistic operating conditions, white Gaussian noise with a standard deviation of $\sigma_{\text{noise}} = 0.1$ is added to the normalized outputs.

The analysis is conducted both in \textit{zero-shot} learning and in \textit{few-shot} adaptation setting. In the zero-shot setting, the pre-trained meta model is directly applied to predict the CSTR output. {In the \textit{few-shot} setting, a lightweight fine-tuning phase is introduced following the methodology in~\cite{piga2023adaptation}, allowing the meta model to adjust to new conditions with minimal computational effort.}

{Two distinct fine-tuning scenarios are investigated:}
\begin{itemize}
    \item {\textbf{Scenario~1: Input adaptation only.} 
    The pre-trained meta model, originally trained on WH systems, is adapted by changing only the excitation signal from multisine to RBS, while keeping the same system class.
    Fine-tuning is performed for 1000 epochs.}
    \item {\textbf{Scenario~2: Input and system adaptation.} 
    Adaptation data are generated by simulating CSTR systems with randomly perturbed physical parameters and using RBS inputs. 
    To mimic a realistic few-shot setting, fine-tuning is carried out over a small set of $25$ batches, each containing $32$ systems, for a total of 1000 training epochs. 
    Although the optimization runs for multiple epochs, the model repeatedly sees the same limited set of systems, making the adaptation effectively few-shot.}
\end{itemize}

{While the second scenario allows adaptation to both the excitation signal and the system class, the first corresponds to a more practical situation in which the user may not know the exact class of dynamical systems but does have control over the excitation signals.}

To evaluate all the cases, 320 distinct CSTR realizations were generated by randomly perturbing selected physical parameters. 
The results, summarized in Table~\ref{tab:out_of_distr}, show that the proposed lightweight fine-tuning procedure markedly improves the generalization capability of the pre-trained meta model. 
In the zero-shot setting, the WH-trained meta model is applied directly to the CSTR without any adaptation, resulting in a relatively large RMSE of $1.249$.
In Scenario~1, we keep the WH training class fixed and adapt only the excitation class from multisine to RBS, which yields a fivefold reduction in RMSE to $0.242$ compared to the zero-shot setting.
In Scenario~2, where both the input and the system class are adapted via CSTR simulations, the RMSE further decreases to $0.115$, which is close to the noise floor $\sigma_{\rm noise}=0.1$.
It is worth noting, however, that the performance achieved in Scenario~1 is already remarkably close to this lower bound, despite adapting only the excitation input while keeping the meta-training system class unchanged.
This indicates that a simple input-matching step can close a substantial portion of the sim-to-real gap, while the remaining gap can be bridged by a few-shot adaptation of the system class.
Overall, the two scenarios highlight that, in certain practical situations, adapting the input representation alone is sufficient to recover most of the performance gains achievable through full adaptation, offering a practical and computationally efficient solution when the system dynamics are unknown but the input signal can be designed by the user.

\begin{table}[!tb]
    \centering
    \caption{CSTR case study: RMSE without and with fine-tuning where only the input class changes (Scenario 1) and where both input and system class change (Scenario 2) during fine-tuning.}
    \small
    \setlength{\tabcolsep}{4pt}
    \begin{tabular}{cccc}
        \toprule
        Setting & Zero-shot & Scenario 1 & Scenario 2 \\
        \midrule
        RMSE & 1.249 & 0.242 & 0.115 \\
        \bottomrule
    \end{tabular}
    \label{tab:out_of_distr}
\end{table}

\begin{figure}[!bt]
\centering
\includegraphics[width=.45\textwidth]{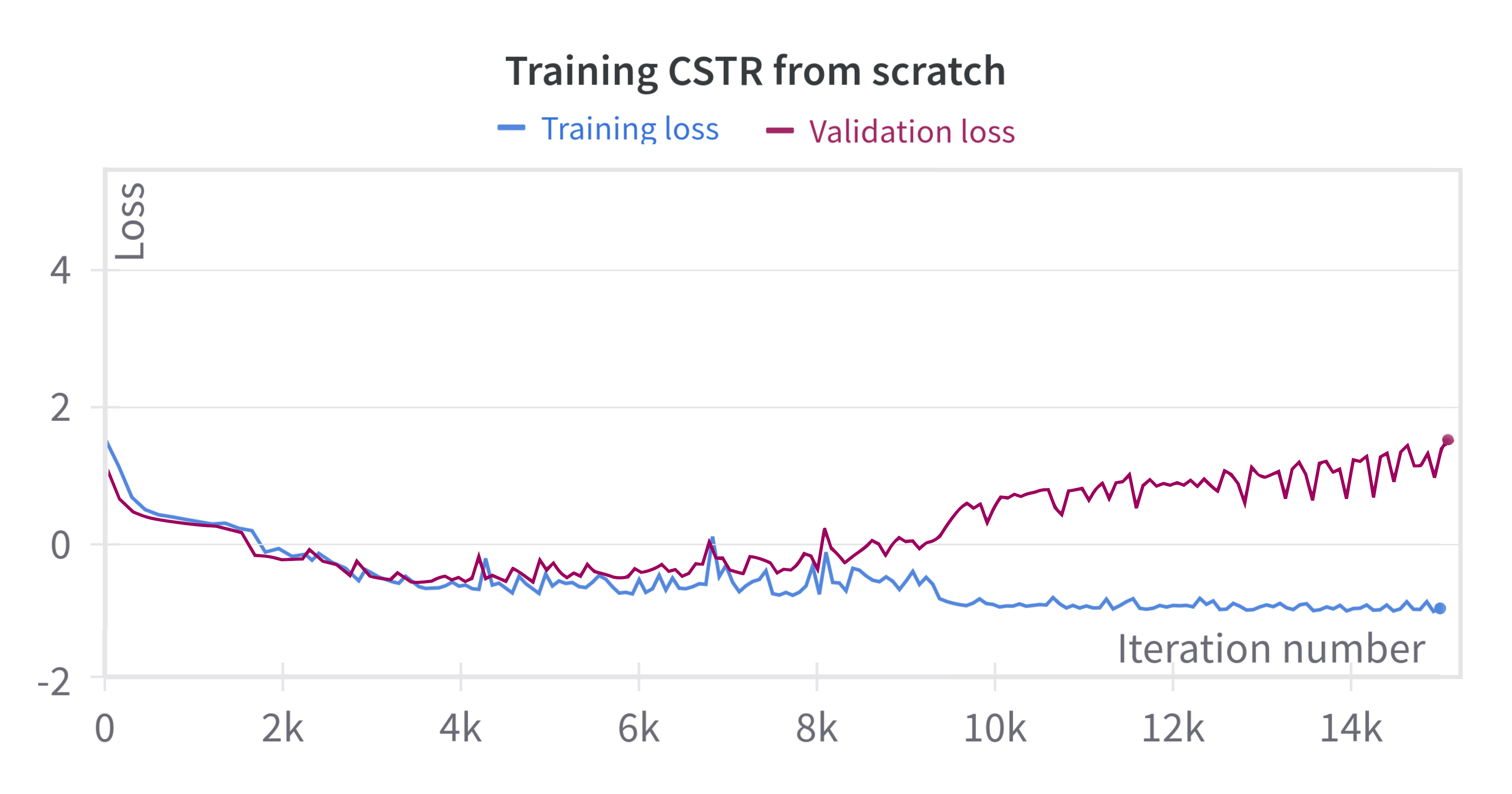}
\caption{{Training and validation curves for the meta model trained from scratch on the CSTR system class. 
The plot shows the evolution of the negative log-likelihood loss over training iterations for both the training (blue) and validation (red) datasets. A divergence between the training and validation losses is observed after approximately 4000 iterations, indicating overfitting. 
 }}

\label{fig:CSTR_scratch}
\end{figure}

For comparison, an additional analysis is conducted, with a meta model  trained from scratch directly on the CSTR system class. To ensure fairness, the training process utilized the same number of system realizations used for fine-tuning previously (namely, $25$ batches of $32$ systems). As shown in Figure~\ref{fig:CSTR_scratch}, this model overfits the small CSTR training set, and the best validation model achieves a test RMSE of 0.136, slightly worse than the fine-tuned meta model. Moreover, training from scratch requires about 4000 iterations to reach its best validation performance, compared to 1000 iterations for fine-tuning. This comparison highlights a practical advantage of the proposed approach: even when the simulator used for pre-training does not perfectly match the target system, a small amount of real or simulator data in the new domain can be used to adapt the pre-trained meta model more efficiently and effectively than training a new meta model from scratch.

\subsection{Synthetic data generation for CED benchmark}

 \begin{figure}[!bt]
\centering
\includegraphics[width=.45\textwidth]{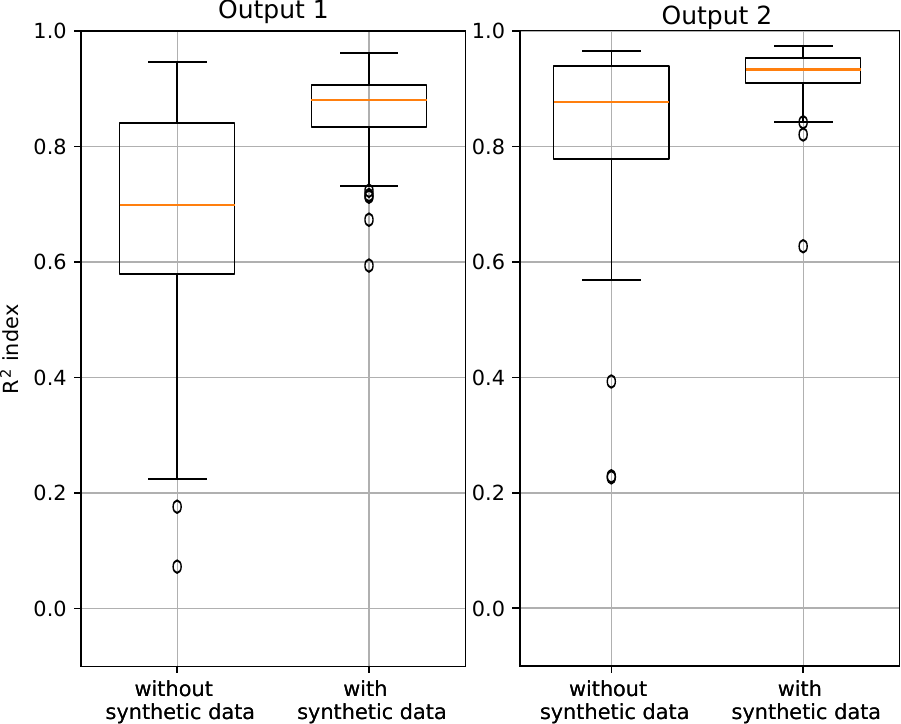}
\caption{{Performance comparison on the CED benchmark when training a dynoNet model with and without synthetic data generated by the pre-trained meta model. 
Boxplots report the distribution of the $R^2$ index over 100 Monte Carlo runs for both benchmark outputs. 
Using only real data (left) leads to higher variance and lower median $R^2$, while augmenting the training set (right) with meta model-generated synthetic trajectories significantly improves  accuracy.}}

\label{fig:synthetic_data_CED}
\end{figure}

In this section, we consider the CED benchmark to demonstrate the capabilities  of the pre-trained meta model in generating synthetic data, as discussed in Section~\ref{sec:synthetic_data_generation}. This benchmark is known to be challenging because of the small size of the training dataset, which consists of only $400$ samples.

Since the deterministic and the maximum-likelihood-based formulations achieved similar results,  we only report the ones achieved with the maximum-likelihood approach.\footnote{Results related to the determinist formulation for synthetic data generation are reported on a simulation example over the WH systems in the conference paper \cite{piga2024synthetic}.}

As a parametric model $\G$ to describe the CED system, a dynoNet neural network~\cite{forgione2021dynonet} is employed. Its architecture comprises 
LTI layers, which process input sequences through filtering operations, combined with memoryless nonlinear components. Training is performed over $6000$ iterations of AdamW. To mitigate overfitting, early stopping is utilized, where the first $300$ samples of the training set are used to train the dynoNet, and the remaining $100$ samples are used for validation. The same validation dataset is also employed to tune the regularization parameter $\lambda$ in the loss function \eqref{eqn:optML}. The same $300$ training samples serve as a context for the pre-trained meta model $\free_\phi$ to generate  additional $200$ synthetic samples, using a multisine signal as a query input. A Monte Carlo simulation with $100$ runs is conducted, where at each run, the initial weights of the dynoNet are randomly initialized, and a new synthetic query input sequence is considered. We remark that the synthetic data was generated by the meta model $\free_\phi$, which was pre-trained on simulated WH models without any further fine-tuning or prior knowledge of the target system.

Results on the test dataset are reported in Figure~\ref{fig:synthetic_data_CED}, which shows a boxplot of the $R^2$ index on the two outputs of the test data over the $100$ Monte Carlo runs, with and without using synthetic data. These results highlight the potential of synthetic data to enhance model robustness and performance, with a median $R^2$ index of $0.70$ on the first output ($0.88$ on the second one) without synthetic data, and a median of $0.88$ on the first output ($0.93$ on the second one) when the synthetic dataset is used. It is also noteworthy to observe the variability of the results across the Monte Carlo runs. Indeed, when no synthetic data is used, variability is solely due to the initial condition of the dynoNet weights. On the other hand, when using synthetic data, variability arises from the initial condition of the dynoNet weights, the selected hyper-parameter $\lambda$, and  the generated synthetic data. Nevertheless, less variability is observed on the achieved $R^2$  when synthetic data is used.

\section{Conclusions} \label{sec:conclusions}

In this paper, we presented novel developments on  in-context learning for system identification, which enables zero-shot prediction of a system of interest without the need to estimate a model for that specific system. This approach eliminates the necessity to implement traditional system identification algorithms, thereby bypassing the need to select hyper-parameters typically chosen by system identification experts, such as model type, model order, training algorithm, etc. Indeed, the meta model is trained offline using a large set of synthetic data and has the capability to directly infer the behavior of the system of interest based on the available context (namely,  available training data) and then predict the system's output for any query input. 

New developments based on sequence patching and ensembling also allow us to process long context sequences, circumventing the quadratic complexity of the attention mechanism of Transformers. Moreover, the meta model can generate synthetic data that can be used for data augmentation in classic system identification. This synthetic data allows to leverage  information from systems in the class, as well as efficient processing of training  data, implicitly learned by the encoder during the pre-training of the meta model.

Numerical results   demonstrate how training a meta model with simulated data allows it to generalize well to real data or systems from a different class, potentially with quick adaptation.

Current research activities are focused on two distinct and seemingly opposite directions. On one hand, we aim at scaling to very large-scale meta models, characterized by billions of parameters. These models are expected to describe very broad classes of dynamic systems, pushing the boundaries of what dynamics meta models can comprehend and predict. On the other hand, research is also focusing on creating distilled architectures. These lighter models aims to enhance computational efficiency and accessibility, making advanced meta models techniques more feasible for real-time applications and edge devices with limited processing capabilities.

\paragraph*{Limitations}

The estimated meta model $\free_\phi$ is able to predict, in a zero-shot learning fashion, the output of a given system of interest based on its context input/output sequence and for any possible query input. However, this predictive meta model also comes with limitations. First, meta training requires a substantial offline computational effort on a large synthetic corpus, which is not needed by classical system identification tools when they are applied to a single system with an ad hoc model structure~\cite{pillonetto2023deep}. Second, the meta model is relatively heavy at inference time, making it less suitable for embedded or resource-constrained implementations compared to parsimonious parametric models. Third, while the meta model achieves competitive accuracy in regimes where per-system data are scarce or weakly informative, classical identification methods trained directly on abundant, high-quality data for a specific system can still reach higher asymptotic accuracy. Finally, the meta model provides predictions rather than an immediately interpretable parametric representation, which can be a drawback when frequency-response analysis, sensitivity studies, or control design require a compact and structured model of the system of interest, although recent works on transformer-based model predictive control~\cite{chen2026successive} indicate that such applications are possible but currently more specialized and computationally demanding.

Further developments in the field are expected to draw inspiration from advancements in Natural Language Generation algorithms, exploring the use of alternative architectures beyond Transformers, like Structured State Space and diffusion models. Additionally, there is potential to leverage open models like LLaMA or DeepSeek, applying low-rank adaptations to their weights to tailor them for the specific meta modeling challenges of dynamical systems.

\bibliographystyle{plain} 
\bibliography{references}        
\appendix 
\section*{Appendix}
\section{Minimum-variance weights for ensembling}

In this appendix section we show that  the variance $\sigma_{\mu_k}^2$, for a specific time step $k$, in \eqref{eq:variance} is minimized by the weights in \eqref{eqn:derweight}, i.e.,
\begin{equation} 
    \omega_{k}^i = \frac{(\sigma_{k}^i)^{-2}}{\sum_{j=1}^s (\sigma_{k}^j)^{-2}}.
\end{equation}

The minimal ensembling uncertainty $\sigma_{\mu_k}^2$ in \eqref{eq:variance} is given by the solution of the the optimization problem:
\begin{equation}
    \label{eq:variance_ap}
   \min_{\omega_k^1,\ldots,\omega_k^s}\sum_{i=1}^s (\omega_{k}^i \sigma_{k}^i)^2. 
\end{equation}
 subject to the constraint $\sum_{i=1}^s \omega_{k}^i = 1$. Using the method of Lagrange multipliers, the augmented Lagrangian is:
 \begin{equation}
    \mathcal{L}(\omega_k, \gamma) = \sum_{i=1}^s (\omega_{k}^i \sigma_k^i)^2 + \gamma\left(\sum_{i=1}^s \omega_{k}^i - 1\right).
\end{equation}

Setting the partial derivatives of $ \mathcal{L}(\omega_k, \gamma)$ to zero, we obtain:
 \begin{align}
    \frac{\partial \mathcal{L}}{\partial \omega_{k}^i} &= 2\omega_{k}^i (\sigma_{k}^i)^2 + \gamma = 0 \implies \omega_{k}^i = \frac{-\gamma}{2(\sigma_{k}^i)^2}, \label{eq:first_partial} \\
    \frac{\partial \mathcal{L}}{\partial \gamma} &= \sum_{i=1}^s \omega_{k}^i - 1 = 0 \implies \gamma = -\sum_{i=1}^s 2(\sigma_{k}^i)^2. \label{eq:second_partial}
\end{align}

Substituting the expression for $\gamma$ from \eqref{eq:second_partial} into \eqref{eq:first_partial}, we obtain the optimal weights in \eqref{eqn:derweight}:
\begin{equation} 
    \omega_{k}^i = \frac{(\sigma_{k}^i)^{-2}}{\sum_{j=1}^s (\sigma_{k}^j)^{-2}}, \quad i = 1, \dots, s.
\end{equation}

\end{document}